\pdfoutput=1

\documentclass[11pt]{article}

\usepackage{nlp4MusA}

\usepackage{times}
\usepackage{latexsym}

\usepackage[T1]{fontenc}

\usepackage[utf8]{inputenc}

\usepackage{microtype}

\usepackage{inconsolata}

\usepackage{graphicx}

\usepackage{times}
\usepackage{latexsym}

\usepackage{color}

\usepackage{lineno}
\usepackage{ascmac}

\title{Can Impressions of Music be Extracted from Thumbnail Images?}

\author{Takashi Harada \\
  The University of Tokyo \\\And
  Takehiro Motomitsu \\
  Hokkaido University \\\And
  Katsuhiko Hayashi \\
  The University of Tokyo \\\AND
  Yusuke Sakai \\
  NAIST \\\And
  Hidetaka Kamigaito \\
  NAIST \\\AND
  \\[-25pt]
  \texttt{katsuhiko-hayashi@g.ecc.u-tokyo.ac.jp} \\}

\begin{document}
\maketitle
\begin{abstract}
In recent years, there has been a notable increase in research on machine learning models for music retrieval and generation systems that are capable of taking natural language sentences as inputs.
However, there is a scarcity of large-scale publicly available datasets, consisting of music data and their corresponding natural language descriptions known as music captions.
In particular, non-musical information such as suitable situations for listening to a track and the emotions elicited upon listening is crucial for describing music.
This type of information is underrepresented in existing music caption datasets due to the challenges associated with extracting it directly from music data.
To address this issue, we propose a method for generating music caption data that incorporates non-musical aspects inferred from music thumbnail images, and validated the effectiveness of our approach through human evaluations.
Additionally, we created a dataset with approximately 360,000 captions containing non-musical aspects.
Leveraging this dataset, we trained a music retrieval model and demonstrated its effectiveness in music retrieval tasks through evaluation.
\end{abstract}

\section{Introduction}
\label{sec:intro}

The enjoyment of music is a highly personal experience, and the ways in which individuals find pleasure in it vary greatly.
One method of enjoying music involves listening to tracks that best suit one's current mood or situations, such as during events like birthdays or in accordance with the season.
Furthermore, individuals often find pleasure in music through discovering songs that match their preferred genres or instruments, and even in creating new music by utilizing the characteristics of existing music.
As our interaction with music becomes more sophisticated, the development of music retrieval and generation systems that are customized to individual tastes and preferences is increasingly important.

\begin{table}
\centering
\resizebox{\linewidth}{!}{%
  \begin{tabular}{lll}
  \\\hline
  \rule{0pt}{2.5ex}
  \vspace{1mm}
   & 
  \bfseries Musical aspects &
  \bfseries Non-musical aspects 
  \\\hline
  \rule{0pt}{9mm}
  \vspace{1.5mm}
  \bfseries Perspective & 
  \begin{tabular}{l}
    - genre \\
    - instruments \\
    - tempo \\
  \end{tabular} &
  \begin{tabular}{l}
    - suitable situations \\
    - seasons, times of day \\
    - evoked emotions \\
  \end{tabular}
  \\\hline
  \rule{0pt}{9mm}
  \vspace{1.5mm}
  \bfseries Music caption  &
  \begin{tabular}{l}
    ``Modern jazz \\
    played with a \\
    triumphal trumpet''
  \end{tabular} &
  \begin{tabular}{l}
    ``Calm song ideal \\
    for listening on a night \\
    with a visible starry sky''
  \end{tabular} \\\hline  
  \end{tabular}
  }
  \caption{Examples of musical and non-musical aspects.}
\label{tab:aspects}
\end{table}
In the domain of music retrieval, studies such as MusCALL~\cite{manco2022contrastive} have investigated the potential for efficient retrieval of music using natural language sentences as inputs, employing contrastive learning methodologies~\cite{le2020contrastive}.
Efforts similar to those of MuLan~\cite{huang2022mulan} are currently being pursued in the field.
In the realm of music generation, research initiatives such as MusicLM~\cite{agostinelli2023musiclm}, MUSICGEN~\cite{copet2023simple}, Mubert~\cite{Mubert-Inc}, and Riffusion~\cite{Riffusion} are integrating elements from systems like MuLan to produce high-quality music from textual descriptions.
These music retrieval and generation models are dependent on natural language descriptive texts, commonly known as music captions, which contain information pertinent to music.

However, a notable challenge emerges, as the bulk of music caption data utilized in these models and systems is not accessible to the public, largely due to copyright constraints.
Furthermore, because existing descriptive texts primarily focus on musical information, they hinder users without extensive musical knowledge from efficiently conducting music retrieval and generation that is based solely on musical elements.
Consequently, there is an urgent need to develop and disseminate music caption data that includes a variety of elements, both musical and non-musical aspects, to the public.

To ensure the diversity of music caption data, it is imperative to include descriptions from two distinct perspectives: musical and non-musical aspects, as delineated in Table~\ref{tab:aspects}.
The musical aspects encompass information like genre and tempo, which can be extracted from music data through music information processing, as exemplified by initiatives such as LP-MusicCaps~\cite{doh2023lpmusiccaps} and others~\cite{liu2023music,Manco_2021}.
In contrast, non-musical aspects include individual impressions and emotional responses associated with music, encompassing elements such as suitable situations for listening to a track, associated seasons and times of day.
The direct extraction of such information from music data poses significant challenges.
Consequently, there is a notable deficiency in music caption data that includes non-musical aspect descriptions, and efforts toward their automatic generation are markedly limited.

To address this issue, we suggest focusing on the thumbnail images associated with music clips on platforms such as YouTube to extract information pertaining to non-musical aspects.
Generally, thumbnail images significantly influence user engagement and click-through rates for content.
In the context of music clips, thumbnail images serve as a succinct visual representation of the music's impression, enabling users to decide at a glance whether it aligns with their music preferences.
In this research, we introduce a methodology for the automatic generation of music caption data, enriched with non-musical aspects derived from these thumbnail images.
The validation results demonstrate that our proposed method is capable of accurately generating music captions that express non-musical aspect information, outperforming baseline methods.
Furthermore, we have made public a dataset~(\url{https://github.com/hu-kvl/llava_music_caption}) containing approximately 360,000 music captions developed using our approach. 
The utility of this dataset was also assessed within a music retrieval model, further substantiating its effectiveness.
The details regarding the dataset creation and the evaluation of the music retrieval model are provided in the Appendices.

\section{Related Work}
To address the scarcity of music caption datasets, the LP-MusicCaps (LLM-Based Pseudo Music Captioning) initiatively utilized GPT-3.5 Turbo~\cite{ChatGPT}, a Large Language Model (LLM), for generating music captions across various tasks employing a comprehensive music tag dataset~\cite{law2009evaluation,inproceedings}.
The dataset produced, encompassing approximately 2.2 million captions and 500,000 audio clips paired with tags, has been made publicly available.
It facilitates the automated generation of music captions with an enriched vocabulary while ensuring alignment with tags and grammatical precision.
Notably, these methods utilize LLMs to focus on generating music captions that incorporate musical elements derived from tags, in contrast to our approach of enriching captions with non-musical aspects.

While our study focuses on the automatic generation of music captions enriched with non-musical aspects using thumbnail images, there exists related research that leverages image intermediary representations for generating lyrics~\cite{watanabe2023text}.
This research tackles the challenge of assisting users in conveying appropriate messages and words in lyric creation.

Contrastive Language-Image Pre-training (CLIP)~\cite{radford2021learning} uses a dual-encoder architecture to embed images and text into a shared latent space.
CLIP has been influential in its demonstration of extracting highly generic visual representations from natural language, and MusCALL~\cite{manco2022contrastive} also adopts this dual-encoder contrastive learning approach pioneered by CLIP.
MuLan~\cite{huang2022mulan} represents a similar endeavor in this field.
Additionally, CLAP~\cite{clap} is a model that applies CLIP's methodology to the audio domain.
A cross-modal retrieval method between music and images has been also investigated in~\cite{nakatsuka2023content}.
In this study, we train a music retrieval model using pairs of music captions and music data created through the proposed method, and assess its effectiveness in music retrieval tasks where input queries are written in natural language sentences.

\begin{figure}[t]
\centering
\begin{center}
\includegraphics[width=8cm]{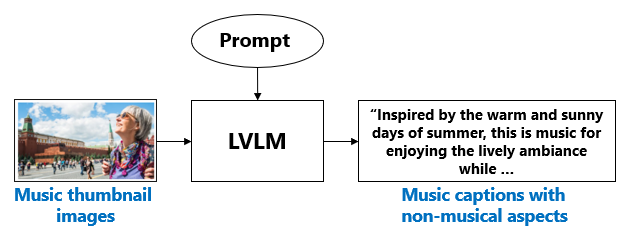}
\end{center}
\caption{Overview of the music captioning method using thumbnail images.}
\label{fig:f1}
\end{figure}

\begin{table*}[t]
\centering
\resizebox{\textwidth}{!}{%
\begin{tabular}{llr}
\hline
Evaluation Metric & Details of Evaluation Metric & Score\\
\hline
Positive & 
\begin{tabular}{@{}l@{}}

Assuming a music generation model that generates music from the provided text, \\
this text contains and accurately expresses the corresponding non-musical aspects.

\end{tabular} 
& 2 \\
\hline
Neutral & 
\begin{tabular}{@{}l@{}}

Assuming a music generation model that generates music from the provided text, \\
this text contains the corresponding non-musical aspects, but their expression is not accurate. 

\end{tabular} 
& 1 \\
\hline
Negative & 
\begin{tabular}{@{}l@{}}
This text does not contain the corresponding non-musical aspects. \\
\end{tabular} 
& 0 \\
\hline
\end{tabular}
}
\caption{Metrics for the human evaluation.}
\label{tab:ta2}
\end{table*}

\begin{figure*}[ht]
\begin{center}
\includegraphics[width=16cm]{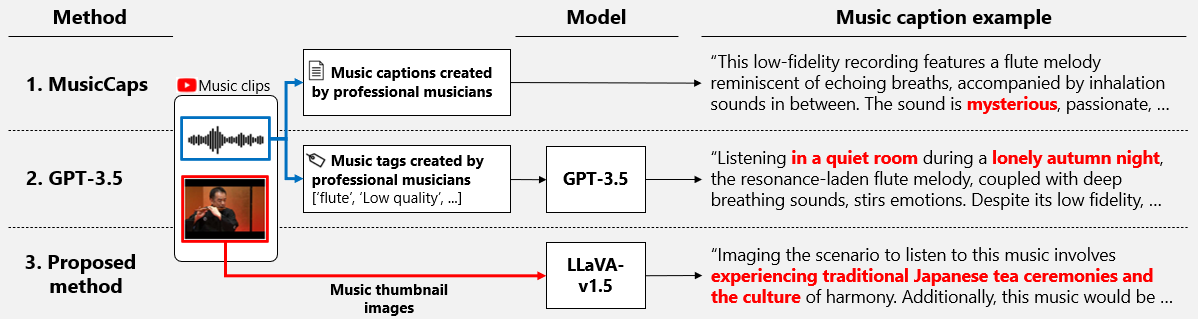}
\end{center}
\caption{Details of the proposed music captioning method and the methods used for comparison.}
\label{fig:f2}
\end{figure*}

\section{Proposed Method}

This study is centered around thumbnail images linked to music clips on platforms like YouTube.
Figure~\ref{fig:f1} provides an overview of our proposed methodology.
In our approach, we first input the thumbnail image into a Large Vision-Language Model (LVLM), a large-scale language model capable of processing images.
Following this, the LVLM, adapted using a meticulously crafted prompt, generates a music caption.
This procedure facilitates the automatic generation of music captions that incorporate non-musical elements.

To ensure that the features of the thumbnail images are not included in the captions generated by the LVLM, we developed a strategic prompting method.
Specifically, we guide the model to initially describe the features of the image itself, such as the content depicted in the thumbnail or its overall mood, in the initial segment of the generated text. 
In the subsequent part, we prompt the model to articulate non-musical aspects that are evoked by these features or the overall mood.

Moreover, by directing the LVLM to distinctly express the features of the thumbnail image in the first portion of the generated text, we successfully separate the description of thumbnail image features from the music caption generation.
This not only achieves effective compartmentalization but also leverages techniques like the Chain of Thought~\cite{wei2022chain}, resulting in the production of high-quality music captions enriched with diverse non-musical elements.

\section{Human Evaluation}
\label{section:human}

\subsection{Experimental Setup}

For the generation of captions to be evaluated, we utilized the MusicCaps~\cite{agostinelli2023musiclm} dataset\footnote{\url{https://huggingface.co/datasets/google/MusicCaps}}.
MusicCaps primarily includes descriptions of music information from YouTube music clips, including corresponding music captions and music tags created by professional musicians.
In our study, we selected 50 songs from MusicCaps, ensuring a balanced distribution across genres, for evaluation.
For each song in the evaluation, three types of captions were prepared:

\begin{itemize}
\setlength{\itemsep}{0pt}
\setlength{\parskip}{0pt}
\item[1.] Manually created captions originally assigned to the dataset (MusicCaps).
\item[2.] Captions automatically generated using a LLM in accordance with the LP-MusicCaps~\cite{doh2023lpmusiccaps} study approach, employing music tags (GPT-3.5).
\item[3.] Captions automatically generated from thumbnail images using the LVLM (Proposed method).
\end{itemize}

The distinct caption generation processes for each method are depicted in Figure \ref{fig:f2}.
For GPT-3.5, we employed the GPT-3.5 Turbo language model, instructing it to describe non-musical aspects based on music tags and those tags themselves.
For the proposed method, we selected the open-source LLaVA-v1.5~\cite{liu2023visual,liu2023improved}\footnote{We performed 4-bit inference on LLaVA-v1.5 13B.} as the LVLM and tuned the prompt to align with our proposed method.
The average lengths of the captions were 148.7 characters for MusicCaps, 146.0 characters for GPT-3.5, and 135.8 characters for the proposed method.

To evaluate each set of captions, a human evaluation was conducted.
For the proposed method, only sections of the generated results related to non-musical aspects, located in the latter part of the output, were extracted for evaluation.

\subsection{Evaluation Procedure and Metrics}

For the human evaluation component, we enlisted two adult Japanese male speakers as evaluators.
These evaluators listened to each music clip for approximately one minute before reviewing the three captions generated by each method.
They then rendered absolute evaluations using a three-point scale (positive, neutral, negative) based on three perspectives (we referred to \cite{manco2022song}): suitable situations for listening to the track (Situation), associated times of day or seasons (Time/Season), and the emotions evoked as a result of listening (Emotion), as delineated in Table \ref{tab:ta2}.
To circumvent order bias, the captions were presented in a randomized sequence.

To facilitate the aggregation of results, the evaluators' three-tier assessments were converted into scores, as illustrated in Table \ref{tab:ta2}.
The final results reflect the average scores across evaluators, providing a cumulative score for the full dataset of 50 entries for each perspective.
Furthermore, the aggregate values for each perspective and the average count of tracks where all three perspectives achieved a perfect score, designated as the ``All 2s count,'' are also reported.
It is predicated on the premise that captions with higher scores and greater counts not only contain non-musical aspects but also exhibit expressiveness.

\begin{table}[t] 
\centering 
\resizebox{\linewidth}{!}
{%
\begin{tabular}{lrrr}
\hline
 & MusicCaps & GPT-3.5 & \textbf{Proposed Method} \\
\hline
Situation & 39.0 & 46.5 & \textbf{75.5} \\
Time/Season & 6.0 & 35.0 & \textbf{72.0} \\
Emotion & 36.5 & \textbf{89.0} & 83.0 \\
\hline
Total & 81.5 & 170.5 & \textbf{230.5} \\
All 2s count & 1.5 & 13.0 & \textbf{23.5} \\
\hline
\end{tabular}%
}
\caption{Human evaluation scores and the count of All 2s.}
\label{tab:ta3}
\end{table}

\subsection{Results and Discussion}

The evaluation results are presented in Table \ref{tab:ta3}, and actual music caption examples are shown in Figure \ref{fig:f2}, using the case of MusicCaps with ``YouTube Video ID (ytID):0u5-WiBKam8'' as an illustration.

Focusing on the aggregate scores across three perspectives, our proposed method achieved the highest evaluation outcomes.
When compared to GPT-3.5, our method surpassed it by a total of 60 points in aggregate scores.
Based on this result, it can be inferred that music captions, including non-musical aspects, can potentially be generated using information extracted from thumbnail images alone, without relying on musical or non-musical tag information created by professional musicians.
Moreover, in every perspective, the scores of both our method and GPT-3.5 exceeded those of manually created captions by humans, MusicCaps.
This suggests that when evaluating music captions based on three perspectives, LVLMs and LLMs can potentially be used to efficiently generate music captions that include non-musical aspects.

Finally, with a focus on the actual music caption examples, Figure \ref{fig:f2} displays instances where non-musical aspects are expressed, written in red letters.
Captions generated by our method are detailed and expressive, providing a rich representation of non-musical aspects.
In contrast, captions by MusicCaps and GPT-3.5 include fewer descriptions of non-musical aspects and are more abstract.

\section{Conclusion}
In this study, we examined the importance of non-musical aspects, such as suitable situations for listening, times of day or seasons, and emotions evoked by listening. Specifically, we addressed the issue of the shortage of music caption data that includes non-musical aspect information, which is essential for constructing music retrieval and generation models. To address this issue, we proposed a method for generating music captions by leveraging thumbnail images. The effectiveness of the proposed method was validated through human evaluations.
It has been reported that using the latest LVLMs to generate descriptions related to artworks has been successful~\cite{hayashi-etal-2024-towards, saito2024evaluating, ozaki2024towards}, and we plan to update our experiments in accordance with the further advancements of LVLMs.


\bibliography{nlp4MusA}

\appendix

\section{Appendices}

\subsection{Open Music Caption Dataset}
\label{section:creation}
In this research, we employed the proposed method to create an open music caption dataset comprising approximately 360,000 pairs of music clips and their associated non-musical aspect captions.
For this dataset, 15 music genres were selected based on their common usage in music databases~\cite{goto2003rwc}, datasets~\cite{defferrard2017fma}, and streaming services.
Table~\ref{tab:data_num} provides a detailed breakdown of each genre along with the corresponding number of data entries.
The music clips used were sourced by searching for the respective genre names and using the tracks found in playlists, with no intentional filtering applied.
However, some clips that could not be acquired due to reasons such as privacy or age restrictions were excluded.

\begin{figure}[t]
\centering
\begin{center}
\includegraphics[width=8cm]{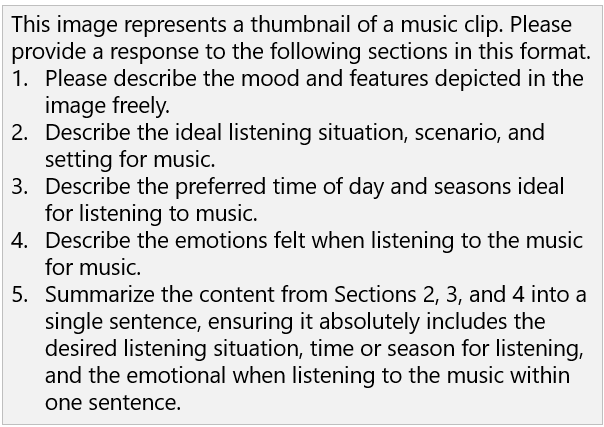}
\end{center}
\caption{Prompt 1: prompt to generate music captions from thumbnail images.}
\label{fig:prompt}
\end{figure}

The construction of the training dataset entailed utilizing the proposed method to input thumbnail images of YouTube music clips and Prompt1 (see Figure~\ref{fig:prompt}) into the LLaVA model.
Prompt1 is designed to facilitate the generation of content across five sections, encompassing the description of the mood and features of the thumbnail image, the preferred listening situation, scenario, and setting, the ideal times of day and seasons for listening, the emotions experienced while listening to the music, and a summarizing sentence that encapsulates the aspects detailed in sections 2, 3, and 4.
The structured prompts were devised to distinctly separate the description of image features from the caption generation process, thereby enhancing the efficacy of the Chain of Thought approach.

The resulting training dataset includes fields such as ``YouTubeID,'' ``URL,'' ``Genre,'' ``Caption,'' and ``Sentence'' for each music clip.
The ``Sentence'' field contains the entire content of Prompt1, enabling its potential use as a comprehensive, non-musical aspect-inclusive music caption.

\begin{table}[t]
\centering
\small
\begin{tabular}{lrrr}
\hline
Genre & Train & Test (All 2s) & Total\\
\hline
house & 49,942 & 80 (49) & 50,022\\
edm & 41,406 & 80 (53) & 41,486\\
classic & 32,695 & 80 (55) & 32,775\\
chill & 31,888 & 80 (62) & 31,968\\
lofi & 27,084 & 80 (62) & 27,164\\
nightcore & 24,717 & 80 (46) & 24,797\\
anime & 24,664 & 80 (55) & 24,744\\
pop & 24,658 & 80 (53) & 24,738\\
rock & 24,425 & 80 (37) & 24,505\\
instrumental & 23,483 & 80 (59) & 23,563\\
tropical house & 21,655 & 80 (56) & 21,735\\
jazz & 12,676 & 80 (51) & 12,756\\
r\&b & 8,258 & 80 (60) & 8,338\\
hiphop & 7,812 & 80 (49) & 7,892\\
bigroom & 5,542 & 80 (43) & 5,622\\
\hline
Total & 360,905 & 1,200 (790) & 362,105\\
\hline
\end{tabular}%
\caption{Number of data samples by genre.}
\label{tab:data_num}
\end{table}

\subsection{Evaluation Dataset}
For the creation of the evaluation dataset, our objective was to generate ground truth data that aligns with human perceptions and sensitivities towards music. 
From the training dataset, we randomly selected 80 music clips for each of the 15 genres, resulting in a total of 1,200 clips for evaluation.
These clips were assessed by a musically knowledgeable adult using a three-point scale based on the perspectives detailed in Section~\ref{section:human}, following approximately a minute of listening to each clip.
The evaluated aspects included the ideal listening situation, the most suitable times of day or seasons for listening, and the emotions experienced during listening.

The evaluation dataset comprises fields such as ``YouTubeID,'' ``URL,'' ``Genre,'' and ``Caption,'' as well as ``Situation,'' ``Time/Season,'' and ``Emotion'' for each music clip.
The latter three fields encapsulate the scores obtained from the human evaluations, corresponding to the three outlined perspectives.
It is crucial to note that both the training and evaluation datasets are entirely in English.
Table~\ref{tab:data_num} details the breakdown and data count for each genre, including the number of music clips that received a ``All 2s.''

\subsection{Validation with music retrieval Model}
\label{section:mr}

To evaluate the appropriateness of the newly created dataset for general user retrieval, we conducted an assessment by training a music retrieval model on the music caption dataset.
The efficacy of the model is gauged based on its proficiency in accurately retrieving relevant items from the dataset.

\subsubsection{Experimental Setup}

To validate the effectiveness of the created dataset, we utilized the MusCALL music retrieval model~\cite{manco2022contrastive}.
MusCALL is a cross-modal contrastive learning model that facilitates both text-based music retrieval and music-based natural language retrieval.
The architecture of the model includes a Transformer~\cite{vaswani2023attention,cliptransformer} for text encoding and a ResNet-50~\cite{he2016powerful}, equipped to handle audio input, for audio encoding.
The embeddings generated by these encoders are utilized to calculate cross-entropy loss, aimed at minimizing the cumulative losses of text and audio embeddings.
The model's configurations, such as hyperparameters, are aligned with those established in MusCALL.

\subsubsection{Dataset}

For the training and evaluation of the MusCALL model, three distinct datasets - train, validation, and test - are employed.
To construct the validation dataset, we randomly selected 30,000 entries from the training data, as delineated in Section~\ref{section:creation}.
Furthermore, 790 entries that are all categorized as ``All 2s'' from the 1,200 human-evaluated music clips were utilized as the test dataset.
A detailed breakdown of the data utilization is provided in Table~\ref{tab:data_num}.

It is crucial to emphasize that the audio data used for this validation consists of 30-second clips extracted from YouTube music clips.
These clips are accompanied by captions that are derived from the non-musical aspects indicated in the dataset's ``Caption'' column.

\subsubsection{Evaluation Metrics}

This experiment fundamentally represents a cross-modal retrieval task, tasked with evaluating the model's proficiency in retrieving items across different modalities. To assess the model's retrieval accuracy, we employ standard cross-modal retrieval metrics, namely Recall at K (R@K) and Median Rank (MedR). Specifically, we set K = \{1, 5, 10\}, which entails measuring the percentage of queries where the correct pair is ranked within the top k and the median rank of the correct pair, respectively, across various genres.

\begin{table}[t]
  \centering
  \small
    \begin{tabular}{lrrrrrr}
      \hline
      Genre & R@1↑ & R@5↑ & R@10↑ & MedR↓ \\
      \hline
      house & 12.2 & 38.8 & \textbf{59.2} & \textbf{8} \\
      anime & 5.5 & 27.3 & \textbf{56.4} & \textbf{9} \\
      instrumental & 16.9 & 37.3 & \textbf{57.6} & \textbf{9} \\
      jazz & 7.8 & 35.3 & 51.0 & 10 \\
      classic & 10.9 & 36.4 & 50.9 & 10 \\
      pop & 5.7 & 20.8 & 47.2 & 11 \\
      rock & 10.8 & 35.1 & 46.0 & 11 \\
      chill & 6.5 & 24.2 & 37.1 & 14 \\
      nightcore & 2.2 & 19.6 & 37.0 & 14 \\
      tropical house & 10.7 & 23.2 & 41.1 & 14 \\
      hip hop & 2.0 & 28.6 & 42.9 & 15 \\
      big room & 0.0 & 9.3 & 25.6 & 17 \\ 
      edm & 7.5 & 17.0 & 32.1 & 17 \\
      lofi & 12.9 & 30.6 & 38.7 & 17 \\
      r\&b & 3.3 & 20.0 & 28.3 & 23 \\
      \hline
      Average & 7.7 & 26.9 & 43.4  &13.3\\
      \hline
    \end{tabular}%
  \caption{Results of text to audio cross-modal retrieval task.}
  \label{tab:text_audio}
\end{table}

\subsubsection{Results and Discussion}

The outcome of the cross-modal retrieval tasks, namely retrieving music from text, is delineated in Table \ref{tab:text_audio} for each genre.
In genres such as house, anime, instrumental, jazz, classic, pop, and rock, the model displays high accuracy in pinpointing relevant pairs, as reflected by the R@10 and MedR metrics in Table \ref{tab:text_audio}.
This underlines the dataset's efficacy for these genres.

Conversely, in genres like big room, hip hop, and R\&B, the model shows diminished retrieval performance, notably in terms of R@1. An examination of the dataset distribution (Table \ref{tab:data_num}) indicates a significant scarcity of data entries in these genres compared to others.
This paucity of data is identified as a key contributor to the reduced retrieval performance.
Therefore, it is imperative to enrich the dataset for genres with limited entries to improve the model’s retrieval proficiency in those specific areas.

\subsubsection{Evaluation as Music Retrieval System}
\begin{table}[t]
\centering
\resizebox{\linewidth}{!}{%
\begin{tabular}{lccp{40mm}r}
\hline
Ranking & YouTubeID & Thumbnail & \multicolumn{1}{c}{Title} & Evaluation \\
\hline
1 & POHHAe0eTfA & \raisebox{-0.5\height}{\includegraphics[width=1.0in]{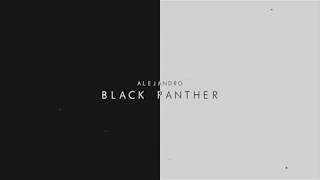}} & ALEJANDRO - Black Panther & Good \\ 
2 & W3H\_8l6dvZs & \raisebox{-0.5\height}{\includegraphics[width=1.0in]{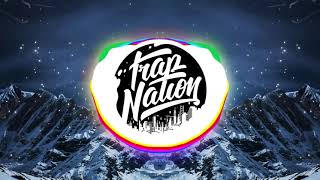}} & Crypto - Faded (feat. Constance) & Fair\\
3 & kFDfvMr1Pd4 & \raisebox{-0.5\height}{\includegraphics[width=1.0in]{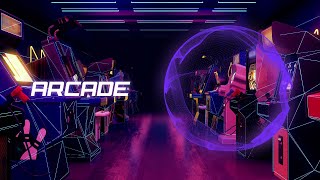}} &  JOXION - 094 [Arcade Release] & Excellent \\ 
4 & N4Up97ZOv0g & \raisebox{-0.5\height}{\includegraphics[width=1.0in]{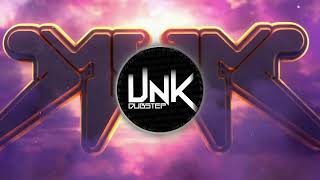}} &  Sora - Changes & Good \\ 
5 & LL1owMMYP78 & \raisebox{-0.5\height}{\includegraphics[width=1.0in]{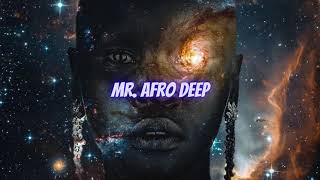}} & Black Gypsies - Kira (Original Mix) & Poor \\ 
6 & AWXsZYhlmwg & \raisebox{-0.5\height}{\includegraphics[width=1.0in]{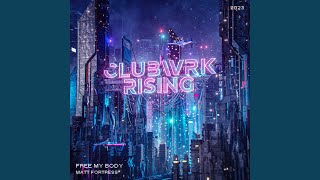}} & Free My Body & Excellent \\ 
7 & 8xn5gH3XUkQ & \raisebox{-0.5\height}{\includegraphics[width=1.0in]{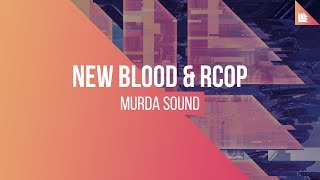}} & New Blood \& RCOP - Murda Sound & Good \\ 
8 & JY97C7EztAg & \raisebox{-0.5\height}{\includegraphics[width=1.0in]{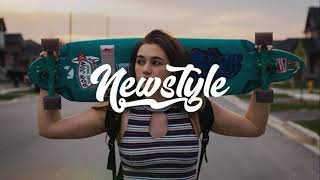}} & S'hustryi Beats - Anti-Covid19 & Poor \\ 
9 & q3ugvrgOri4 & \raisebox{-0.5\height}{\includegraphics[width=1.0in]{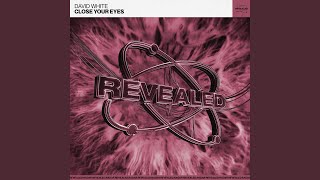}} &  Close Your Eyes (Extended Mix) & Good \\
 
\hline
\end{tabular}
}
\caption{Retrieval results and evaluations for the query ``Energetic songs for a late-night drive.''}
\label{tab:energetic}
\end{table}

In this section, we conduct a more pragmatic evaluation of the experiments.
Our aim is to validate the effectiveness of the dataset developed in this study for constructing a music retrieval system that incorporates non-musical aspects in retrieval queries.

Applying the ranking methodology utilized in the cross-modal retrieval task, we implemented a retrieval system that is capable of retrieving music from text queries. This retrieval was executed across all 790 music clips from various genres in the evaluation dataset, each with All 2s. A text query was employed: ``Energetic songs for a late-night drive.'' The top 9 results for the query were presented, accompanied by the author's subjective evaluations of the music clips, categorized as ``Excellent,'' ``Good,'' ``Fair,'' or ``Poor.''

The findings are exhibited in Table \ref{tab:energetic}. Listening to the music clips retrieved for ``Energetic songs for a late-night drive,'' the top result resonates with the energetic and intense vibe suitable for a nighttime drive.
This outcome indicates the dataset's capability in supporting a music retrieval system based on non-musical aspects.

Furthermore, despite the lack of explicit musical aspect information in both queries, the retrieval system successfully retrieved relevant music. This implies that the system can cater to users without specific music-related expertise, enabling music retrieval based on impressions and emotional cues, thereby enhancing the diversity and flexibility of music retrieval.

Therefore, the retrieval system evaluated in this analysis demonstrates its proficiency in retrieving music appropriate for non-musical aspect descriptions in natural language queries. The results underscore the efficacy of the dataset created in this study for developing a comprehensive music retrieval system.

\end{document}